%% file: PKMTerminology_Corrected_Arxiv.tex
\documentclass[10pt,twocolumn]{asme2ej}
\usepackage{amsfonts}
\usepackage{color}
\usepackage{amsmath}

\usepackage{amssymb}
\usepackage{graphicx}

\input{tcilatex}
\begin{document}

\input{opening_PKMTerminology_revised.inp}

\section{Introduction}

Redundancy has been introduced to provide the dexterity needed for
manipulation tasks and to overcome the kinematic and dynamic limitations of
PKM. Kinematically redundant PKM were first proposed as variable geometry
trusses (VGT) \cite{MiuraFuruya1985},\cite{PadmanabhanArunReinholtz1992},%
\cite{ReinholzGokhale1987} where special attention was payed to the elastic
properties \cite{JungCraneRoberts2006},\cite{Seguchi1990}. Although several
design concepts were proposed, like tetrahedron-based VGT \cite%
{JainKramer1990}, 2 degree of freedom (DOF) planar PKM \cite%
{EbrahimiCarreteroBoudreau2008}, and hyper-redundant PKM \cite%
{Chirikjian1995}, kinematically redundant PKM did not draw as much attention
as redundantly actuated PKM (RA-PKM) did. Moreover redundant actuation
schemes were developed to increase and homogenize the kinematic
manipulability and stiffness, to increase the achievable acceleration, and
to eliminate singularities and thus enlarge the usable workspace, as
reported for a number of prototypes in \cite{Abedinnasab},\cite{buttolo},%
\cite{cheng2003},\cite{GoguDETC2007},\cite{kock98},\cite{Saglia},\cite%
{Wang2009}, \cite{Wang2009Robotica},\cite{ZhaoGao2009},\cite{ZhaoGao2011},
and references in \cite{merlet96}. Actuation redundancy can be achieved by
actuation of passive joints or by additional kinematic chains connecting the
moving and fixed platform, which is the standard approach. It is also
beneficial for the calibration as shown in \cite{jeong},\cite{Yiu2003},\cite%
{ZhangCongShangLiJiang2007} since it provides additional sensor data. It was
shown in \cite{obrien}, \cite{zhang} that actuation redundancy improves the
kinematic manipulability and eventually avoids singularities. The latter
were investigated in \cite{firmani},\cite{liao}. Stiffness and force
capabilities were studied in \cite{chakarov2004},\cite{dasgupta},\cite%
{nokleby}. Investigations of optimal force distribution and the related
design issues have been reported in \cite{kim},\cite{kurtz},\cite{LeePark},%
\cite{lee},\cite{nahon}. The advantage of redundantly actuated PKM are,
however, owed to a more complex control, in particular if model-based
controlled is envisioned, as pointed out in \cite{cheng2003},\cite{gardner},%
\cite{hufnageliftomm},\cite{tro2005},\cite{icra2011},\cite{iftommPKM},\cite%
{Saglia},\cite{ZhaoGao2009}. One critical point is the redundancy resolution
within the feedforward part of the controller. Another challenge is the
presence of undesired antagonistic control forces, that are due to geometric
model uncertainties \cite{tro2010} but are also inherent in decentralized
control schemes \cite{icra2011}. They cause elastic deformations that must
be taken into account for the calibration in particular of heavy payload
RA-PKM \cite{Ecorchard}.

Despite the advances in control and design of redundant PKM there is yet no
consistent terminology, however. Whereas for redundant serial manipulators
the terminology was clarified already in \cite{conkur} there is no
established classification of redundant PKM. In particular actuation
redundancy is frequently referred to as overactuation, and occasionally
redundantly actuated PKM are confused with overconstrained mechanisms, but
clear definitions and concepts are essential for the systematic design and
analysis of novel RA-PKM.

This paper aims on clarifying the terminology for redundant PKM in general
and RA-PKM in particular. As basis for such a classification, a kinematic
model for PKM is first introduced in section \ref{kinematics}, and the PKM
motion equations are recalled in section \ref{dynamics} that are used in the
model-based control admitting the representation of PKM as non-linear
control system in section \ref{control}. As preparation, singularities of
PKM are classification as input, output, and c-space singularities in
section \ref{sec:Critical}. This is used in section \ref{secModes} to
distinguish different actuation modes. Section \ref{secKinRedundancy}
recalls the definition of kinematic redundancy adopted to PKM. The main
contribution of this paper is the definition of actuation in section \ref%
{secActuation} upon the non-linear control system. PKM are further
classified as full-actuated and underactuated.

The discussion explicitly refers to the geometric aspects of redundancy. The
central object in the kinematics of PKM is the c-space in which the
manipulator motion is encoded. In contrast to serial manipulators, where the
c-space is a smooth manifold, its geometry is usually rather complicated for
PKM, and it is only locally a smooth manifold. The abundance of
singularities in the c-space is reflected by non-smooth motions impairing
the PKM control. On the other hand it is this c-space geometry, and its
dimension relative to the number of actuators, that gives rise to different
actuation schemes. From a geometric point of view the actuation corresponds
to a parameterization of the c-space in terms of actuator coordinates, and
the c-space of a PKM being embedded in the joint space allows for redundant
actuation of PKM by combining different parameterizations. It is pointed out
geometrically that input singularities can be avoided by redundant actuation.

\section{%
%TCIMACRO{\TeXButton{\label{sec:modeling}}{\label{sec:modeling}} }%
%BeginExpansion
\label{sec:modeling}
%EndExpansion
Parallel Manipulator Modeling}

\subsection{Manipulator Kinematics%
%TCIMACRO{\TeXButton{\label{kinematics}}{\label{kinematics}}}%
%BeginExpansion
\label{kinematics}%
%EndExpansion
}

A PKM consists of a moving platform, carrying an end-effector (EE), that is
connected to the base platform by several (possibly identical) kinematic
chains (limbs, struts, legs) containing actuated joints. PKM are thus
characterized by closed kinematic loops imposing certain constraints. Denote
with $\mathbf{q}\in \mathbb{V}^n$ the vector of joint variables $q^a,{%
a=1,\ldots ,n}$ (higher DOF joints are split into 1 DOF joints), where $%
\mathbb{V}^n:=\mathbb{T}^{n_R}\times \mathbb{R}^{n_P}$ is the joint space of
a PKM comprising $n_R$ revolute and $n_P$ prismatic/screw joints. The vector 
$\mathbf{q}\in \mathbb{V}^n$ is referred to as the configuration of the PKM.
A configuration is clearly admissible only if it satisfies the loop
constraints. The $r$ geometric constraints are summarized in the system 
\begin{equation}
\mathbf{0}=\mathbf{h}\left( \mathbf{q}\right) ,\;\mathbf{h}\left( \mathbf{q}%
\right) \in {\mathbb{R}}^r.  \label{geomconstr}
\end{equation}
Time differentiation yields the kinematic constraints 
\begin{equation}
\mathbf{0}=\mathbf{J}\left( \mathbf{q}\right) {\dot{\mathbf{q}}},\;\mathbf{J}%
\left( \mathbf{q}\right) \in {\mathbb{R}}^{r,n}  \label{constr1}
\end{equation}
with the constraint Jacobian $\mathbf{J}$.

In any practical implementation there are further constraints due to joint
limits or to prevent collisions. This is expressed as a system of $c$
inequality constraints 
\begin{equation}
\mathbf{0}\leq \mathbf{g}\left( \mathbf{q}\right) ,\;\mathbf{g}\left( 
\mathbf{q}\right) \in {\mathbb{R}}^c.  \label{g}
\end{equation}
Then the configuration space of the PKM, being the set of all admissible
configurations, is 
\begin{equation}
V:=\left\{ \mathbf{q}\in {\mathbb{V}}^n|\mathbf{h}\left( \mathbf{q}\right) =%
\mathbf{0},\mathbf{g}\left( \mathbf{q}\right) \geq \mathbf{0}\right\} .
\label{Cspace}
\end{equation}
It is instructive to consider the c-space geometry. Neglecting inequality
constraints (\ref{g}), $V$ is a variety and only locally (i.e. closed to a
given configuration $\mathbf{q}$) a smooth manifold. These manifolds are
separated by the singular points of $V$, where the rank of $\mathbf{J}$
changes (section \ref{sec:CSing}). If in the neighborhood of $\mathbf{q}$ in 
$V$ the number of locally independent constraints is $\mathrm{rank}\,\mathbf{%
J}\leq r$, the local DOF of the PKM is $\delta _{\text{loc}}:=n-\mathrm{rank}%
\,\mathbf{J}$. It is important to notice that the DOF is in fact a local
property of the mechanism, and that the PKM might even attain different
mobilities without disassembling it as for so-called kinematotropic
mechanisms \cite{Wohlhart1996}. The maximal local DOF is referred to as the
global DOF denoted with $\delta $.

The PKM interacts with its environment via an EE --the mechanical output.
This EE is represented by an EE-frame that is rigidly attached to it. The
configuration of this EE-frame relative to a world-fixed (inertial) frame is
represented by a matrix $\mathbf{C}\in SE\left( 3\right) $ \cite{murray}.
The \textbf{output mapping} $f_{\text{O}}:\mathbb{V}^n\rightarrow SE\left(
3\right) $, yields the EE-configuration $\mathbf{C}=f_{\text{O}}\left( 
\mathbf{q}\right) $ in terms of the configuration $\mathbf{q}$. Then the 
\textbf{workspace} of the PKM is the set of attainable EE-configurations: 
\begin{equation}
\mathcal{W}:=\{f_{\text{O}}\left( \mathbf{q}\right) |\mathbf{q}\in
V\}\subset SE\left( 3\right) .
\end{equation}
Usually only a part of this $\mathcal{W}$ is used depending on the
application and the presence of singularities.

The EE-velocity is represented by a twist coordinate vector $\widehat{%
\mathbf{V}}\in se\left( 3\right) $, respectively $\mathbf{V}\in {\mathbb{R}}%
^6$. The instantaneous EE kinematics is determined by 
\begin{equation}
\mathbf{V}=\mathbf{J}_{\text{O}}\left( \mathbf{q}\right) \dot{\mathbf{q}}
\label{JE}
\end{equation}
relating the EE-velocities to the state of the PKM, where $\mathbf{J}_{\text{%
O}}\left( \mathbf{q}\right) :T_{\mathbf{q}}\mathbb{V}^n\rightarrow se\left(
3\right) $ is the \textbf{output Jacobian}.

The PKM motion is determined by the motion of its actuators --the mechanical
inputs. The relation of actuator and PKM motion is expressed by the \textbf{%
input mapping} $f_{\text{I}}:\mathbb{V}^n\rightarrow \mathcal{I}$ that
assigns to any PKM configuration the admissible inputs. This relation may
not be unique as there may be different inputs corresponding to the same
configuration. If the PKM is equipped with $m$ actuator inputs, $\mathcal{I}$
is $m$-dimensional. The inputs are not necessarily the motion of some joints
of the PKM since the actuators may act at arbitrary locations of the PKM,
e.g. via tendons or pulleys. Generally arbitrary exogenous inputs could be
considered. In the following it is assumed that $m$ joints are directly
actuated. Then, the joint coordinate vector can be split into coordinates of 
$m$ active and $n-m$ passive joints, $\mathbf{q}_{\text{a}}$ and $\mathbf{q}%
_{\text{p}}$, respectively. That is, $\mathbf{q}_{\text{a}}$ are the
kinematic inputs, and the input mapping is simply the projection of $\mathbb{%
V}^n$ to the corresponding $m$-dimensional subspace. With this splitting the
kinematic constraints (\ref{constr1}) become 
\begin{equation}
\mathbf{0}=\mathbf{J}_{\text{p}}\left( \mathbf{q}\right) \dot{\mathbf{q}}_{%
\text{p}}+\mathbf{J}_{\text{a}}\left( \mathbf{q}\right) \dot{\mathbf{q}}_{%
\text{a}}  \label{constr2}
\end{equation}
where $\mathbf{J}_{\text{p}}\left( \mathbf{q}\right) \in \mathbb{R}^{r,n-m}$%
, $\mathbf{J}_{\text{a}}\left( \mathbf{q}\right) \in \mathbb{R}^{r,m}$. 
\begin{figure*}[tbh]
{\includegraphics[width=17cm]{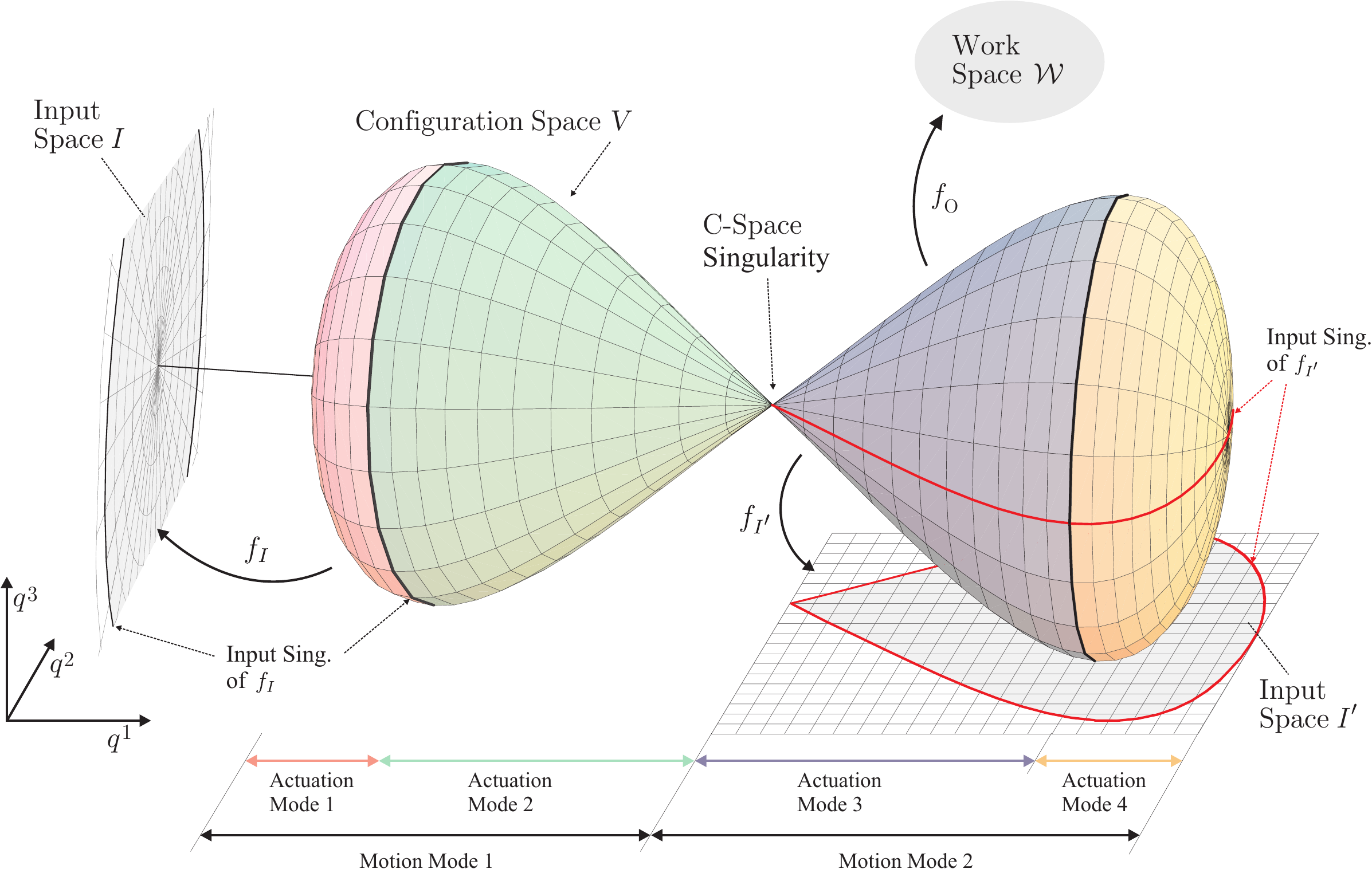}}
\caption{Geometric interpretation of the PKM control system. The operation
modes refer to input space $\mathcal{I}$}
\label{figScheme}
\end{figure*}

The kinematic PKM model can be schematically represented as 
\begin{equation}
\fbox{$%
\begin{array}{ccccccc}
&  &  &  &  &  &  \\ 
\;\;\;\;\;\; & \mathcal{W} & \overset{f_{\text{O}}}{\longleftarrow } & V & 
\overset{f_{\text{I}}}{\longrightarrow } & \mathcal{I} & \;\;\;\;\;\; \\ 
&  &  &  &  &  & 
\end{array}
$}  \label{model}
\end{equation}
Clearly the central object is the c-space $V$ geometrically representing the
mechanism. The input and output mapping yields the input and output,
respectively, that corresponds to a given configuration. They are not one to
one for PKM in general and for redundant PKM in particular.

Figure \ref{figScheme} shows a schematic representation of this model for a
2-dimensional c-space in a 3-dimensional joint space that can be locally
parameterized by $q^2$ and $q^3$, or by $q^1$ and $q^2$, for instance. The
latter may be used as inputs, and the projection of $V$ on the respective
coordinate subspace gives the input spaces $\mathcal{I}$ and $\mathcal{I}%
^{\prime }$.

The model (\ref{model}) is rather abstract. For the analysis of a PKM in a
given configuration the following instantaneous kinematics model is used 
\vspace{-3ex}
\begin{equation}
\fbox{$%
\begin{array}{crllc}
&  &  &  &  \\ 
& \mathbf{0} &  & =\mathbf{J}_{\text{p}}\left( \mathbf{q}\right) \dot{%
\mathbf{q}}_{\text{p}}+\mathbf{J}_{\text{a}}\left( \mathbf{q}\right) \dot{%
\mathbf{q}}_{\text{a}} &  \\ 
\;\;\;\;\;\;\;\;\;\;\; & \mathbf{V} &  & =\mathbf{J}_{\text{O}}\left( 
\mathbf{q}\right) \dot{\mathbf{q}} & 
\begin{array}{c}
\; \\ 
\;%
\end{array}
\;\;\;\;\;\; \\ 
& \dot{\mathbf{q}}_{\text{a}} &  & =\mathbf{J}_{\text{I}}\dot{\mathbf{q}} & 
\\ 
&  &  &  & 
\end{array}
$}  \label{instmod}
\end{equation}
The first implicit equation locally describes the c-space $V$. The second
equation is the differential output mapping relating instantaneous PKM and
EE motion. The third equation is the differential input mapping that yields
the instantaneous input motion. Since the input mapping is the projection
from $V$ to the $\mathbf{q}_{\text{a}}$ coordinate subspace $I$, the
Jacobian $\mathbf{J}_{\text{I}}$ is constant with entries 1 and 0.

\begin{remark}
%TCIMACRO{\TeXButton{\label{remModel}}{\label{remModel}} }
%BeginExpansion
\label{remModel}%
%EndExpansion
%TCIMACRO{\TeXButton{\rm}{\rm}}
%BeginExpansion
\rm%
%EndExpansion
The instantaneous model (\ref{instmod}) applies to general PKM. It admits to
separately investigate the mechanism's kinematics and its interaction with
the environment via inputs and outputs. It further allows to exhaustively
analyze and classify the corresponding critical phenomena as it is presented
in \cite{zlatanovclassify}. On the other hand the input-output kinematics of
the PKM is often represented in the form $\mathbf{M}_2\mathbf{V}=\mathbf{M}_1%
\dot{\mathbf{q}}_{\text{a}}$ \cite{gosselin2},\cite{tsai}, which is easily
obtained with the reciprocal screw approach. If $\mathbf{M}_2$ is square,
the forward Jacobian is $\mathbf{J}_{\text{F}}=\mathbf{M}_2^{-1}\mathbf{M}_1$%
. That is, the three mappings in the model (\ref{instmod}) are resolved.
However, the internal state of the PKM is hidden.
\end{remark}
\vspace{-2ex}

\subsection{Motion Equations in Minimal Coordinates%
%TCIMACRO{\TeXButton{\label{dynamics}}{\label{dynamics}}}%
%BeginExpansion
\label{dynamics}%
%EndExpansion
}

A dynamical model is indispensable for the model-based control of PKM. A PKM
is a force controlled multibody system (MBS) subject to geometric
constraints due to kinematic loops. In applications where the manipulator
interacts with its environment, the PKM is subject to additional, possibly
non-holonomic, constraints. The latter will not be taken into account here.
There are several approaches for deriving motion equations of constrained
MBS that have different numerical efficiencies. Independently from the
applied principle a basic fundamental difference is the choice of
independent generalized coordinates. It is crucial that the dynamic PKM
model, that is eventually used for the control, is given in terms of a
minimal set of $\delta $ generalized coordinates. The solution of the
geometric constraints (\ref{geomconstr}) can (locally) be expressed in terms
of a subset $\mathbf{q}_2$ comprising $\delta _{\text{loc}}$ joint
variables, so that the configuration is given as $\mathbf{q}=\psi \left( 
\mathbf{q}_2\right) $. In other words, $\mathbf{q}_2$ are local coordinates
on $V$ for a parameterization $\psi $, that admits to describe the internal
kinematics of the mechanism. It is usually impossible to explicitly derive
the relation of $\mathbf{q}$ on some chosen $\mathbf{q}_2$. Therefore, a
standard method for MBS with kinematic loops is to introduce a relation on
velocity and acceleration level after a coordinate partitioning in dependent
and independent joint variables \cite{Amirouche},\cite{Nikravesh1985}. With
this approach computationally efficient forms of PKM motion equations can be
derived as it was pursued in \cite%
{cheng2003,liu2003,tro2005,nakamura,Heimann}.

Denote with $\mathbf{c}\equiv \left( c_1,\ldots ,c_m\right) $ the vector of $%
m$ generalized control forces in the actuated joints, and $\mathbf{\tau }\in
se^{*}\left( 3\right) $ the EE wrench due to external loads at the EE. Then
the PKM dynamics is governed by the motion equations \cite{tro2005}%
%TCIMACRO{\TeXButton{-1.5ex}{\vspace{-1.5ex}}}%
%BeginExpansion
\vspace{-1.5ex}%
%EndExpansion
\begin{equation}
\overline{\mathbf{G}}\left( \mathbf{q}\right) \ddot{\mathbf{q}}_2+\overline{%
\mathbf{C}}\left( \mathbf{q},\dot{\mathbf{q}}\right) \dot{\mathbf{q}}_2+%
\overline{\mathbf{Q}}\left( \mathbf{q},\dot{\mathbf{q}},t\right) +\overline{%
\mathbf{J}}_{\text{E}}^T\left( \mathbf{q}\right) \mathbf{\tau }=\mathbf{A}%
^T\left( \mathbf{q}\right) \mathbf{c},  \label{woronjetz}
\end{equation}
where $\overline{\mathbf{G}}$ is the generalized mass matrix, $\overline{%
\mathbf{C}}\dot{\mathbf{q}}_2$ represents generalized Coriolis and
centrifugal forces, $\overline{\mathbf{Q}}$ represents all remaining
generalized forces. The $m\times \delta _{\text{loc}}$ control matrix $%
\mathbf{A}$ is such that $\dot{\mathbf{q}}_{\text{a}}=\mathbf{A}\dot{\mathbf{%
q}}_2$, and the right hand side $\overline{\mathbf{Q}}_{\text{a}}=\mathbf{A}%
^T\mathbf{c}$ in (\ref{woronjetz}) are the generalized control forces due to
actuator forces $\mathbf{c}$. (\ref{woronjetz}) is a system of $\delta _{%
\text{loc}}$ ordinary differential equations in $\mathbf{q}\in \mathbb{V}^n$
that, together with the $r$ kinematic constraints, completely determines the
PKM dynamics.

If the number $m$ of actuated joints exceeds the PKM DOF (i.e. the PKM is
redundantly actuated), $\mathbf{A}^T$ is not square and has a null-space of
dimension $m-\delta _{\text{loc}}$. Only the actuator forces $\mathbf{c}$
not in the null-space of $\mathbf{A}^T$ are effective control forces. It is
a peculiarity of redundantly actuated PKM that actuator forces can be
generated in the null-space of $\mathbf{A}^T$. Such null-space forces,
giving rise to internal prestress, can be exploited for second level tasks
such as backlash avoidance or stiffness control \cite{kock2010},\cite%
{tro2005},\cite{icra2006},\cite{ShinLeeInJeongKim2011},\cite{yi}.\vspace{-4ex%
}

\section{%
%TCIMACRO{\TeXButton{\label{control}}{\label{control}} }%
%BeginExpansion
\label{control}
%EndExpansion
The Associated Nonlinear Control Systems}

A PKM is a force controlled (holonomically) constrained dynamical system,
whose dynamics is governed by (\ref{woronjetz}). The control purpose is to
manipulate the EE that embodies the system's mechanical output. This makes
the PKM a second order control-affine control system on the configuration
space $V$ that is represented as first order control system \cite{bullo},%
\cite{nijmeijer}%
%TCIMACRO{\TeXButton{-1ex}{\vspace{-1ex}}}%
%BeginExpansion
\vspace{-1ex}%
%EndExpansion
\begin{eqnarray}
\dot{\mathbf{x}} &=&\mathbf{f}\left( \mathbf{x}\right) +\sum_{i=1}^m\mathbf{g%
}_i\left( \mathbf{x}\right) c^i%
%TCIMACRO{\TeXButton{-3}{\vspace{-3mm}} }%
%BeginExpansion
\vspace{-3mm}
%EndExpansion
\label{ctrl2} \\
\mathbf{C} &=&f_{\text{O}}\left( \mathbf{x}\right) 
%TCIMACRO{\TeXButton{-6}{\vspace{-6mm}} }%
%BeginExpansion
\vspace{-6mm}
%EndExpansion
\nonumber
\end{eqnarray}
with state vector $\mathbf{x}:=\left( \mathbf{q}_2,\dot{\mathbf{q}}_2\right) 
$. Therein%
%TCIMACRO{\TeXButton{-1ex}{\vspace{-1ex}} }%
%BeginExpansion
\vspace{-1ex}
%EndExpansion
\begin{equation}
\mathbf{f}:=\;\left( 
\begin{array}{c}
\dot{\mathbf{q}}_2%
%TCIMACRO{\TeXButton{-}{\vspace{0.5mm}} }%
%BeginExpansion
\vspace{0.5mm}
%EndExpansion
\\ 
-\overline{\mathbf{G}}^{-1}%
%TCIMACRO{\TeXButton{-}{\hspace{-0.3mm}} }%
%BeginExpansion
\hspace{-0.3mm}
%EndExpansion
%TCIMACRO{\TeXButton{\big}{\big} }%
%BeginExpansion
\big
%EndExpansion
(\overline{\mathbf{C}}\dot{\mathbf{q}}_2+\overline{\mathbf{Q}}+\overline{%
\mathbf{J}}_{\text{E}}^T\mathbf{\tau }%
%TCIMACRO{\TeXButton{\big}{\big} }%
%BeginExpansion
\big
%EndExpansion
)%
\end{array}
\right) ,%
%TCIMACRO{\TeXButton{-3}{\vspace{-3mm}}}%
%BeginExpansion
\vspace{-3mm}%
%EndExpansion
\end{equation}
is the drift vector field, and the columns $\mathbf{g}_i,i=1,\ldots ,m\leq n$
of%
%TCIMACRO{\TeXButton{-3}{\vspace{-3mm}}}%
%BeginExpansion
\vspace{-3mm}%
%EndExpansion
\begin{equation}
\mathbf{g}:=\left( 
\begin{array}{c}
\mathbf{0}%
%TCIMACRO{\TeXButton{-}{\vspace{0.5mm}} }%
%BeginExpansion
\vspace{0.5mm}
%EndExpansion
\\ 
\overline{\mathbf{G}}^{-1}\mathbf{A}^T%
\end{array}
\right) 
%TCIMACRO{\TeXButton{-1}{\vspace{-3mm}} }%
%BeginExpansion
\vspace{-3mm}
%EndExpansion
\label{InpVecField}
\end{equation}
are the control vector fields, via which the control forces affect the
system. The actuation of the PKM determines the immediate effect of control
forces in a given pose of the PKM. Apparently the degree of actuation has to
do with the number of independent control vector fields, but also with the
vector space spanned by $\mathbf{g}_i$ (see section \ref{secTerminology}).%
\vspace{-3ex}

\section{Critical Configurations%
%TCIMACRO{\TeXButton{\label{sec:Critical}}{\label{sec:Critical}}}%
%BeginExpansion
\label{sec:Critical}%
%EndExpansion
}

Singularities are manifested in a qualitative change of the manipulator's
kinematic and static properties. A PKM may become structurally unstable,
lose the ability to properly interact with its environment, or become
non-manipulable. This must be taken into account for a proper definition of
actuation. For this purpose it is sufficient to distinguish c-space
singularities, output singularities, and input singularities. These
singularity types can occur simultaneously, and their combination, if
possible, may lead to phenomena such as instantaneously impossible input
motions, or instantaneously redundant inputs. An exhaustive study is
presented in \cite{zlatanovclassify}, where six different types are
identified and all possible combinations are listed. In the following
c-space, input, and output singularities are classified as far as necessary
for introducing a sensible notion of redundancy. It is instructive to
explicitly refer to the c-space geometry since this allows interpreting the
different singularities and the actuation redundancy geometrically.
Moreover, the c-space topology reveals all motion characteristics as shown
in \cite{liu2003} and \cite{ARK2008}. In \cite{liu2003} the input, output,
and c-space singularities where analyzed using differential forms. The local
structure of the c-space was addressed in \cite{ARK2008}.

At this point a remark is in order. Singularities are identified upon the
rank of certain Jacobians. It should be stressed that a singular point is
one in which the rank of the considered Jacobian changes, but frequently any
situation is referred to as singular whenever the rank is lower then
expected.

\vspace{-3ex}

\subsection{C-space Singularities%
%TCIMACRO{\TeXButton{\label{sec:CSing}}{\label{sec:CSing}}}%
%BeginExpansion
\label{sec:CSing}%
%EndExpansion
}

The configuration space $V$ is a variety in $\mathbb{V}^n$. $V$ comprises
several connected smooth manifolds (subspaces like smooth curves or
surfaces) that are separated by singular points, indicating 'non-smoothness'
of $V$ at these points. Points of $V$ that belong to a smooth manifold are
called regular. The attribute 'singular', meaning solitary or unique,
reflects the fact that they are special in the sense that almost all points
are regular. Clearly the mechanism's mobility has to do with the dimension
of $V$. A motion of the PKM corresponds to a curve in $V$, and in points
where $V$ is not a smooth manifold, the motion can be non-smooth.

The PKM mobility can be clearly defined upon the c-space topology. The 
\textbf{differential} \textbf{DOF} (or instantaneous DOF) of the mechanism
is defined as $\delta _{\text{diff}}\left( \mathbf{q}\right) :=${$n$}${-%
\mathrm{rank}\,\mathbf{J}(}${$\mathbf{q}$}${)}$. A point $\mathbf{q}\in V$
is \textbf{regular} if and only if it belongs to a submanifold of $V$, i.e.
there is a neighborhood $U\left( \mathbf{q}\right) $ such that ${\delta _{%
\text{diff}}}$ is constant in $U\left( \mathbf{q}\right) \cap V$, otherwise
it is \textbf{singular}. The \textbf{local} \textbf{DOF} in $\mathbf{q}$,
denoted $\delta _{\text{loc}}\left( \mathbf{q}\right) $, is the local
dimension of $V$. This is the highest dimension of manifolds passing through 
$\mathbf{q}$. If $\mathbf{q}$ is regular, then $V$ is locally a $\delta _{%
\text{loc}}\left( \mathbf{q}\right) $-dimensional manifold. In case of
kinematotropic mechanisms \cite{Wohlhart1996} there are different local DOF
in a connected component of $V$. The \textbf{global DOF} $\delta $ is the
highest local dimension of $V$. If $V$ is not connected, i.e. there are
different assemblies of the mechanism, which can not be attained via an
admissible finite motion, the global DOF needs to be restricted to the
relevant assembly. A detailed discussion of the geometric mobility concept
can be found in \cite{ARK2008},\cite{MMTGenMob}.

Since $V$ is locally a smooth manifold if and only if the constraint
Jacobian ${\mathbf{J}}$ has constant (not necessarily full) rank, one can
introduce the following.

\begin{definition}
%TCIMACRO{\TeXButton{\rm}{\rm} }
%BeginExpansion
\rm%
%EndExpansion
A point $\mathbf{q}\in V$ is called a \textbf{c-space singularity}, if and
only if ${\mathrm{rank}\,\mathbf{J}}$ is not constant in any neighborhood of 
$\mathbf{q}$ in $V$.
\end{definition}

As a practical consequence, even if the c-space has locally a dimension $%
\delta _{\text{loc}}$, in a c-space singularity no subset of $\delta _{\text{%
loc}}$ joint variables can be used to parameterize $V$, and the generalized
mass matrix $\overline{\mathbf{G}}$ in (\ref{woronjetz}) is not regular.

For example, the c-space $V\in \mathbb{R}^3$ in figure \ref{figScheme} is a
two-dimensional smooth manifold except at the indicated c-space singularity.
At any other point it is smooth with $\mathbf{J}$ having constant rank, and
a unique tangent plane can be attached to $V$. At the c-space singularity
there is no unique tangent plane, and the differential DOF of the PKM
increases.

\subsection{Input Singularities}

Naturally, a configuration is regarded as an input singularity if the
interdependence of actuator motion and the motion of the PKM undergoes a
qualitative change. To state this more precisely it is necessary to consider
the motion of actuated and passive joints.

\begin{definition}
%TCIMACRO{\TeXButton{\rm}{\rm} }
%BeginExpansion
\rm%
%EndExpansion
The configuration $\mathbf{q}\in V$ is called \textbf{passive singularity} (%
\textbf{actuator singularity}) if $\mathrm{rank}\,\mathbf{J}_{\text{p}}$ ($%
\mathrm{rank}\,\mathbf{J}_{\text{a}}$) is not constant in any neighborhood
of $\mathbf{q}$ in $V$. If $\mathbf{q}$ is either a passive or an actuator
singularity, it is called \textbf{input singularity}.
\end{definition}

C-space singularities are necessarily active and passive, and thus input
singularities (since rank$\,\mathbf{J}$ drops). On the other hand active and
passive singularities can occur simultaneously without being c-space
singularities.

If, in the model in figure \ref{figScheme} the projection of $V$ onto the $%
q^2$-$q^3$ coordinate plane is used as input space $\mathcal{I}$, its
boundary points are passive (input) singularities, since there $V$ is normal
to $\mathcal{I}$ so that instantaneous motions $q^1$ are possible
independently from the inputs. To cope with this problem, in such
configurations, joints 1 and 2 could be used as actuators with input space $%
\mathcal{I}^{\prime }$. Also this actuation scheme exhibits input
singularities at the boundary of $\mathcal{I}^{\prime }$.

The constraints (\ref{constr2}) can be solved for the velocities of passive
and actuator joints. The general solution is respectively 
%TCIMACRO{\TeXButton{-3}{\vspace{-3mm}}}%
%BeginExpansion
\vspace{-3mm}%
%EndExpansion
\begin{eqnarray}
\dot{\mathbf{q}}_{\text{p}} &=&-\mathbf{J}_{\text{p}}^{+}\mathbf{J}_{\text{a}%
}\dot{\mathbf{q}}_{\text{a}}+\dot{\mathbf{q}}_{\text{p}0},\;\;\text{with\ \ }%
\dot{\mathbf{q}}_{\text{p}0}\in N\left( \mathbf{J}_{\text{p}}\right) \\
\dot{\mathbf{q}}_{\text{a}} &=&-\mathbf{J}_{\text{a}}^{+}\mathbf{J}_{\text{p}%
}\dot{\mathbf{q}}_{\text{p}}+\dot{\mathbf{q}}_{\text{a}0},\;\;\text{with\ \ }%
\dot{\mathbf{q}}_{\text{a}0}\in N\left( \mathbf{J}_{\text{a}}\right) , 
\nonumber
\end{eqnarray}
where $N\left( \mathbf{J}\right) $\thinspace is the \thinspace null-space of 
$\mathbf{J}$. The left-pseudoinverse $\mathbf{J}^{+}$ always exists. If $%
\mathrm{rank}\,\mathbf{J}_{\text{a}}<m$, the null-space $N\left( \mathbf{J}_{%
\text{a}}\right) $ is non-empty, and there exist instantaneous motions of
actuator joints even if all passive joints are locked. If $\mathrm{rank}\,%
\mathbf{J}_{\text{p}}<n-m$, the null-space $N\left( \mathbf{J}_{\text{p}%
}\right) $ is non-empty, and there exist instantaneous motions of passive
joints for locked actuator joints. Whether these instantaneous motions
correspond to finite motions depends on whether the considered configuration
is regular, that is whether the rank of the respective Jacobian is constant
in a neighborhood of $\mathbf{q}$. In input singularities the rank of $%
\mathbf{J}_{\text{p}}$ or $\mathbf{J}_{\text{a}}$ drops and the respective
null-space increases.

\begin{remark}
%TCIMACRO{\TeXButton{\rm}{\rm} }
%BeginExpansion
\rm%
%EndExpansion
%TCIMACRO{\TeXButton{\label{rem:EEsing}}{\label{rem:EEsing}} }
%BeginExpansion
\label{rem:EEsing}%
%EndExpansion
In \cite{gosselin2} a classification of input and output singularities of
PKM was proposed upon a formulation of the form $\mathbf{M}_2\mathbf{V}=%
\mathbf{M}_1\dot{\mathbf{q}}_{\text{a}}$ (see also remark \ref{remModel}).
Accordingly, singularities of type I and II are identified when respectively 
$\mathbf{M}_1$ or $\mathbf{M}_2$ is not full rank. Type I and II
singularities are also called serial and parallel singularities,
respectively, as type II only occurs for PKM. Type II singularities are also
termed force singularities, since certain EE-wrenches can not be
equilibrated by control forces. This classification is useful when
considering the PKM as transmission device, relating input and output
motions. The internal state of the PKM is hidden, however. E.g., the PKM may
be in a passive singularity even if both matrices are regular.%
%TCIMACRO{
%\TeXButton{\setcounter{example}{0}}{\setcounter{example}{0}
%}}
%BeginExpansion
\setcounter{example}{0}
%
%EndExpansion
\end{remark}

\begin{figure}[t]
\centerline{\includegraphics[width=8.4cm]{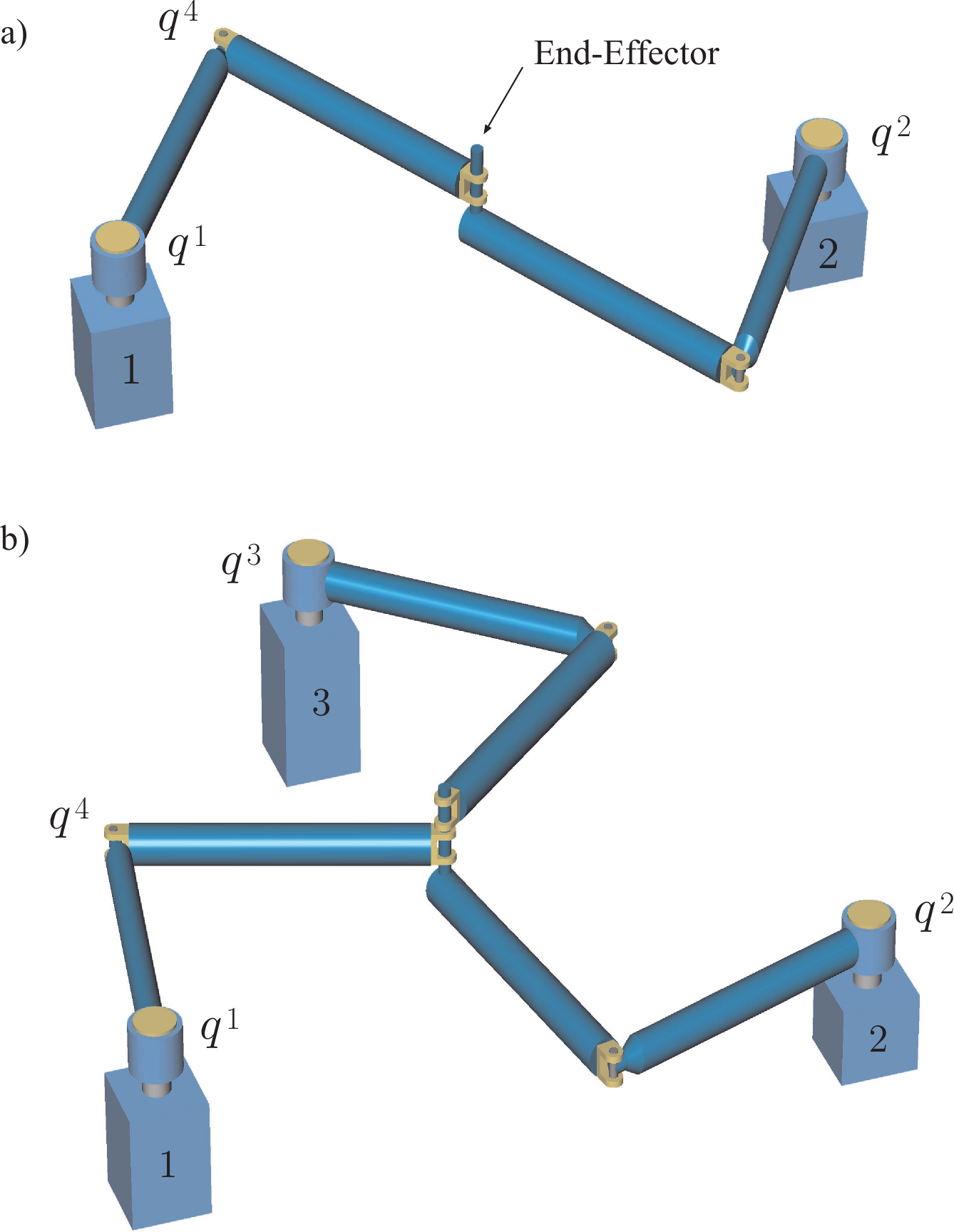}}
\caption{5-bar mechanism a), and its redundant extension b) the \protect%
\underline{R}R/2\protect\underline{R}RR PKM.}
\label{figPantograph}
\end{figure}

\begin{figure}[t]
\includegraphics[width=8.2cm]{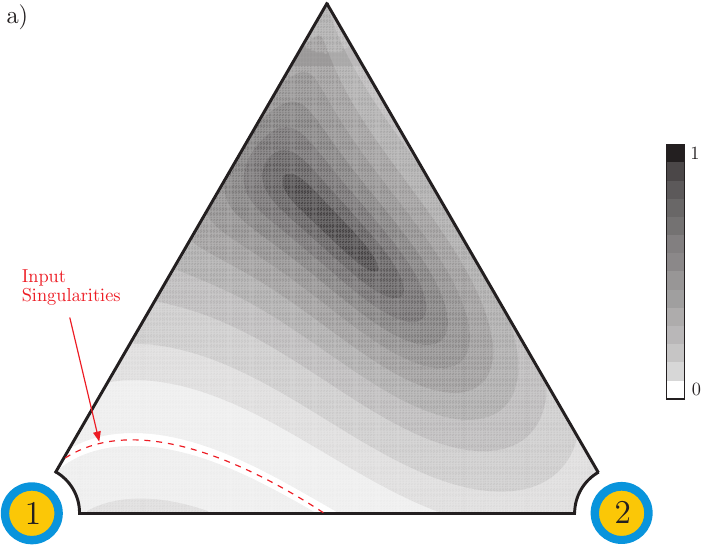} \bigskip %
\includegraphics[width=8.2cm]{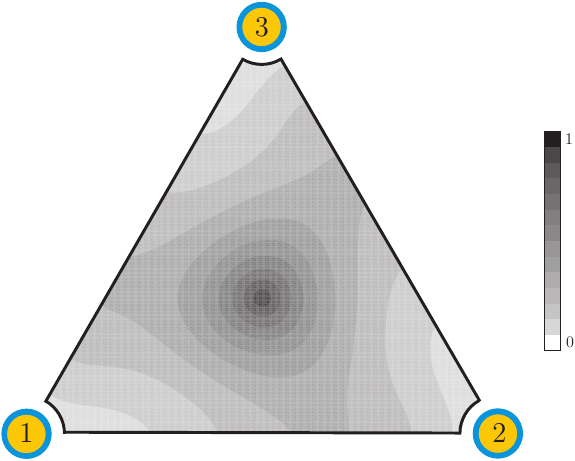}
\caption{Manipulability measure $1/\protect\kappa $ of a) 5-bar mechanism
and b) \protect\underline{R}R/2\protect\underline{R}RR PKM.}
\label{figManip}
\end{figure}

\begin{figure}[h!]
\centerline{\includegraphics[width=8.9cm]{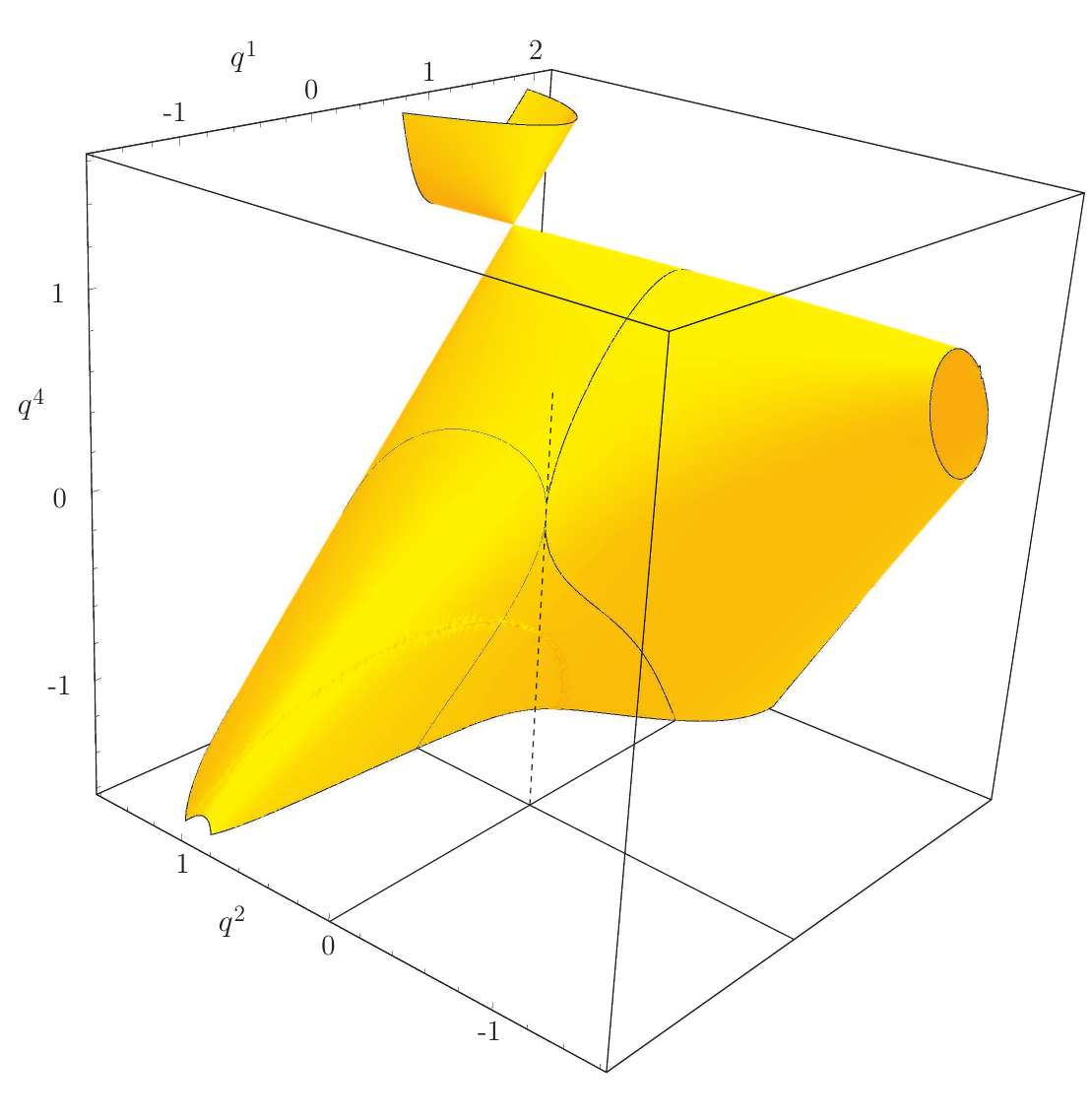}}
\caption{$q^1$-$q^2$-$q^4$ c-space section of the 5-bar mechanism.}
\label{figCSpacePant}
\end{figure}

\begin{example}
%TCIMACRO{\TeXButton{\rm}{\rm} }
%BeginExpansion
\rm%
%EndExpansion
%TCIMACRO{\TeXButton{\label{examPantograph}}{\label{examPantograph}}}
%BeginExpansion
\label{examPantograph}%
%EndExpansion

The 5-bar mechanism in figure \ref{figPantograph}a) and the mechanism in
figure \ref{figPantograph}b), which is obtained by adding a third kinematic
chain between the EE and the base, are examples showing the avoidance of
input singularities by means of redundant actuation. The EE of both
mechanisms can be positioned in the plane, and the two translation
components are the mechanical outputs. The 5-bar mechanism is controlled by
the two drive units at the base. Adding an identical actuated chain to the
5-bar mechanism does not change the DOF nor the EE mobility, so that both
mechanisms have the DOF 2. In the pose shown in figure \ref{figPantograph}a)
the 5-bar mechanism exhibits a passive input singularity. This is revealed
by the manipulability measure defined as the inverse of the condition number 
$\kappa $ of $\mathbf{J}_{\text{F}}\mathbf{J}_{\text{F}}^T$ \cite{murray},%
\cite{parkkim},\cite{yoshikawa1}, where the forward kinematic Jacobian $%
\mathbf{J}_{\text{F}}$ relates the EE velocity $\mathbf{V}$ and actuator
velocities ${\dot{\mathbf{q}}}_{\text{a}}$ according to $\mathbf{V}=\mathbf{J%
}_{\text{F}}{\dot{\mathbf{q}}}_{\text{a}}$ in remark \ref{remModel} (notice
that for this positioning device the measure does not depend on the scaling
of translations and rotations, which is problematic in general for spatial
manipulators). Figure \ref{figManip} shows the distribution of $1/\kappa $
for the EE positions in the workspace. It is clearly visible that the
kinematic dexterity measures of the two manipulators differ significantly.
Moreover, the 5-bar mechanism exhibits singularities, where the EE motion is
not controllable by the actuators, reflected by a drop of manipulability.
The configuration in figure \ref{figPantograph}a) is such an input
singularity. This indeterminacy is removed with the redundant third actuated
chain in figure \ref{figPantograph}b). Following the conventional notation
the mechanism in figure \ref{figPantograph}b) will be denoted \underline{R}%
R/2\underline{R}RR indicating that the EE is connected to the fixed base by
one kinematic chain (chain 2) comprising two revolute joints and two chains
(chains 1 and 3) with three revolute joints, where underlines specify the
actuated joints.\newline
Geometrically the occurrence of an input singularity means that, in the
configuration in figure \ref{figPantograph}a), the 2-dimensional c-space
cannot be parameterized by the two actuator joint angles $q^1$ and $q^2$.
Figure \ref{figCSpacePant} shows the $q^1$-$q^2$-$q^4$ section of the
c-space, where the origin $\mathbf{q}_0=\mathbf{0}$ is assigned to the input
singularity. Apparently the projection of $V$ onto the $q^1$-$q^2$
coordinate plane is not unique, and at $\mathbf{q}_0$ the joint angle $q^4$
does not depend uniquely on the input coordinates $q^1$ and $q^2$. These
input singularities occur whenever the two middle links are parallel, i.e.
when the mechanism resembles a 4-bar, and the EE position for these input
singularities lie on the coupler curve of this 4-bar mechanism, restricted
to the work space as indicated in figure \ref{figManip}a). The
non-uniqueness problem, and thus the input singularities, are removed by
adding a third actuated chain, that yields the redundantly actuated 
\underline{R}R/2\underline{R}RR PKM in figure \ref{figPantograph}b). Then
the mobility is unaltered but the number of parameters used to prescribe the
motion is increased. Beside the increased and homogenized manipulability the
apparent advantage of this redundant actuation is the elimination of input
singularities. The dimension of the joint space is increased without
increasing the dimension of the c-space (i.e. DOF), and the c-space of the 
\underline{R}R/2\underline{R}RR is embedded in a 8-dimensional joint space
(while that of the 5-bar is embedded in a 5-dimensional space), which gives
more freedom for choosing actuator coordinates, or even using a redundant
set. In other words two 5-bar mechanisms are connected and two copies of the
c-pace in figure \ref{figCSpacePant} are glued together by identifying the
respective $q^1$-$q^4$ subspaces.
\end{example}

\subsection{Output Singularities}

The output mapping $f_{\text{O}}$ assigns to any configuration an EE-pose,
and the output Jacobian the EE-twist to the PKM state. Output singularities
are situations where the number of instantaneous motions that are determined
by the PKM motion changes.

\begin{definition}
%TCIMACRO{\TeXButton{\rm}{\rm} }
%BeginExpansion
\rm%
%EndExpansion
The configuration $\mathbf{q}\in V$ is called \textbf{output singularity} if 
$\mathrm{rank}\,\mathbf{J}_{\text{O}}$ is not constant in any neighborhood
of $\mathbf{q}$ in $V$.
\end{definition}

Apparently the occurrence of output singularities depends on how the outputs
are assigned to the PKM, according to $f_{\text{O}}$. They indicate a change
of the way the PKM interacts with its environment, but not critical
configurations of the PKM itself.

\section{%
%TCIMACRO{\TeXButton{\label{secTerminology}}{\label{secTerminology}} }%
%BeginExpansion
\label{secTerminology}
%EndExpansion
Proposal for a Terminology}

\subsection{Operation Modes%
%TCIMACRO{\TeXButton{\label{secModes}}{\label{secModes}}}%
%BeginExpansion
\label{secModes}%
%EndExpansion
}

Critical configurations impair the integrity of the PKM and the stability of
its dynamics model. A reliable operation is only ensured in regular
configurations. The submanifolds of regular points constitute modes of
operation of the PKM. In this regard only those are relevant that can be
attained by a motion starting from the initial assembly, but not those that
could be attained by opening kinematic loops and assembling it differently.

\begin{definition}
%TCIMACRO{\TeXButton{\rm}{\rm} }
%BeginExpansion
\rm%
%EndExpansion
%TCIMACRO{\TeXButton{\label{defMode}}{\label{defMode}}}
%BeginExpansion
\label{defMode}%
%EndExpansion
The connected subvarieties of $V$ are called the \textbf{assembly modes} of
the PKM. The connected submanifolds of regular points of $V$ are called 
\textbf{motion} \textbf{modes} of the PKM. The submanifolds of the motion
modes consisting of configurations that are not input singularities are
called \textbf{actuation modes} of the PKM. The submanifolds of the motion
modes consisting of configurations that are neither input nor output
singularities are called \textbf{operation modes} of the PKM.
\end{definition}

This is a refinement of the c-space according to critical configurations.
The motion modes consists of all regular configurations in which the PKM is
stable in the sense that it does not exhibit c-space singularities. The
actuation modes are the submanifolds where in addition a continuous control
of the PKM is ensured. Finally a further restriction to the configurations
where a continuous interaction with the environment is ensured, so that the
PKM can be operated as transmission device, yields the operation modes. The
motion and actuation modes are indicated for the geometric model in figure %
\ref{figScheme}, which has only one assembly mode.

\begin{remark}[Aspects]
%TCIMACRO{\TeXButton{\rm}{\rm}}
%BeginExpansion
\rm%
%EndExpansion
The motion modes consist of all configurations in which the PKM performs
smooths motions. The actuation modes are submanifolds of the motion modes
that additionally ensure uninterrupted actuation. For serial manipulators
exists an equivalent to operation modes, called aspects, which refer to the
manifolds of regular points of the forward kinematics \cite{borell}.
So-called 'generalized aspects' were introduced for fully-parallel PKM in 
\cite{genaspects} as the connected submanifolds where both Jacobians $%
\mathbf{M}_1$ and $\mathbf{M}_2$ in remark \ref{remModel} are regular. These
submanifolds may, however, comprise other (e.g. c-space) singularities
since, as already mentioned, the internal state of the PKM is ignored.
\end{remark}

It is sensible to classify the tasks of a PKM according to the critical
configurations. To this end consider a task with corresponding task space $%
T\subset SE\left( 3\right) $, i.e. set of poses the EE has to attain. It is
assumed that $T$ is connected.

\begin{definition}
%TCIMACRO{\TeXButton{\rm}{\rm} }
%BeginExpansion
\rm%
%EndExpansion
A task with task space $T$ is a \textbf{regular task}, if there is an
operation mode $M\subset V$ such that $T\subseteq f_{\text{O}}(M)$.
\end{definition}

\begin{remark}
%TCIMACRO{\TeXButton{\rm}{\rm} }
%BeginExpansion
\rm%
%EndExpansion
A regular task can be accomplished without passing through any critical
configuration. An interesting concept in this regard is the notion of
cuspidal serial manipulators \cite{wenger3} that is being adopted for PKM 
\cite{genaspects},\cite{Chablat2011}.
\end{remark}

\subsection{Kinematic Redundancy%
%TCIMACRO{\TeXButton{\label{secKinRedundancy}}{\label{secKinRedundancy}}}%
%BeginExpansion
\label{secKinRedundancy}%
%EndExpansion
}

Traditionally kinematic redundancy refers to situations were the DOF of a
manipulator exceeds the required EE-mobility. This notion can be directly
adopted for PKM.

\begin{definition}
%TCIMACRO{\TeXButton{\rm}{\rm} }
%BeginExpansion
\rm%
%EndExpansion
Consider a PKM with global mobility $\delta $. The PKM is called \textbf{%
kinematically redundant} if $\dim \mathcal{W}<\delta $. The \textbf{degree
of kinematic redundancy} is $\rho _k:=\delta -\dim \mathcal{W}$. The motion
that the PKM can perform with fixed EE-configuration $\mathbf{C}\in SE\left(
3\right) $ is called the \textbf{self motion} with $\mathbf{C}$. The
submanifolds of $S_{\mathbf{C}}:=\{\mathbf{q}|\mathbf{C}=f_{\text{E}}\left( 
\mathbf{q}\right) \}\subset V$ are called the self motion manifolds for this
EE-pose. For serial manipulators this was studied in \cite{Burdick}.
\end{definition}

\begin{remark}
%TCIMACRO{\TeXButton{\rm}{\rm} }
%BeginExpansion
\rm%
%EndExpansion
This definition appears similar to the definition of kinematic redundancy of
serial manipulators. It is important, however, to notice the specifics of
PKM. While the c-space of serial manipulators is a smooth manifold, the
c-space $V$ of a PKM comprises manifolds (possibly of different dimensions,
as for kinematotropic mechanisms) that are separated by c-space
singularities. For this reason the global DOF appears in the above
definition, and it should be noticed that possibly $\dim \mathcal{W}<\delta
_{\text{loc}}$ in one motion mode while $\dim \mathcal{W}\geq \delta _{\text{%
loc}}$ in another mode.
\end{remark}

This notion of kinematic redundancy is solely based on the local dimensions
of c-space and workspace, but does not take into account how (or even if)
the task motion is embedded in the workspace. Even if the PKM is
kinematically redundant it may not be able to accomplish a particular task.
Moreover, the motion characteristics may be different in different motion
modes. For instance the EE may perform planar motions in one mode and
spherical motions in another mode. Therefore it is appropriate to introduce
the notion of task redundancy.

Consider a task with corresponding task space $T\subset SE\left( 3\right) $,
assumed to be connected.

\begin{definition}
%TCIMACRO{\TeXButton{\rm}{\rm} }
%BeginExpansion
\rm%
%EndExpansion
The PKM is called \textbf{task redundant} if $\dim T<\dim \mathcal{W}$ and $%
T\subseteq \mathcal{W}$, and \textbf{task deficient} if $\color{red}W\subset
\mathcal{T}$.
\end{definition}

\begin{remark}
%TCIMACRO{\TeXButton{\rm}{\rm} }
%BeginExpansion
\rm%
%EndExpansion
Notice that a manipulator may be kinematically redundant as well as task
deficient.
\end{remark}

\subsection{Types of Actuation%
%TCIMACRO{\TeXButton{\label{secActuation}}{\label{secActuation}}}%
%BeginExpansion
\label{secActuation}%
%EndExpansion
}

The above preliminaries admit to introduce a stringent terminology for
redundantly actuated PKM. The following definitions refer to a configuration 
$\mathbf{q}$ in a certain actuation mode where the PKM has local DOF $\delta
_{\text{loc}}$.

\begin{definition}
%TCIMACRO{\TeXButton{\rm}{\rm} }
%BeginExpansion
\rm%
%EndExpansion
%TCIMACRO{\TeXButton{\label{defDOA}}{\label{defDOA}}}
%BeginExpansion
\label{defDOA}%
%EndExpansion
In the considered actuation mode, the \textbf{degree} \textbf{of} \textbf{%
actuation} (DOA) is the number of independent input vector fields in the
control system (\ref{ctrl2}):%
%TCIMACRO{\TeXButton{-3}{\vspace{-3mm}}}
%BeginExpansion
\vspace{-3mm}%
%EndExpansion
\begin{equation}
\alpha \left( \mathbf{q}\right) :=\mathrm{rank\,}\mathbf{g}\left( \mathbf{q}%
\right) =\mathrm{rank\,}\mathbf{A}\left( \mathbf{q}\right) .%
%TCIMACRO{\TeXButton{-3}{\vspace{-3mm}} }
%BeginExpansion
\vspace{-3mm}%
%EndExpansion
\label{alpha}
\end{equation}
If $\alpha <\delta _{\text{loc}}$, the PKM is called \textbf{underactuated},
and if $\alpha =\delta _{\text{loc}}$, the PKM is \textbf{full-actuated}.
The \textbf{degree} \textbf{of} \textbf{redundancy} of the actuation is $%
\rho _\alpha :=m-\alpha $. The PKM is called \textbf{redundantly} \textbf{%
actuated} if $\rho _\alpha >0$ and \textbf{non-redundantly} \textbf{actuated}
if {$\rho _\alpha =0$}.
\end{definition}

\begin{remark}
%TCIMACRO{\TeXButton{\rm}{\rm} }
%BeginExpansion
\rm%
%EndExpansion
Redundantly actuated PKM are occasionally termed 'overactuated'.
Notwithstanding that redundantly actuated PKM can be underactuated, a
full-actuated PKM is completely actuated, and an improvement is impossible.
Hence the term 'overactuation' should not be used. Geometrically,
underactuation refers to situations where the active joint variables do not
constitute local coordinates on $V$, i.e. $\mathbf{q}_{\text{a}}$ does not
fully determine the PKM configuration.
\end{remark}

\begin{remark}
%TCIMACRO{\TeXButton{\rm}{\rm} }
%BeginExpansion
\rm%
%EndExpansion
Actuation refers to the immediate effect of control forces on the state of
the system. It is a pointwise property. The effect of the actuation on the
PKM motion is described by the controllability of the system. This is a
local property, i.e. considering the effect of actuation over a small time 
\cite{bullo},\cite{nijmeijer}. Clearly, an underactuated PKM can be
controllable (but it is questionable whether such a PKM offers sufficient
stability).
\end{remark}

\begin{remark}
%TCIMACRO{\TeXButton{\rm}{\rm} }
%BeginExpansion
\rm%
%EndExpansion
The above DOA definition makes explicit use of the expression of (\ref
{InpVecField}) and thus of (\ref{woronjetz}). This is only valid at regular
configurations (i.e. in actuation modes) since in c-space or input
singularities the projected mass matrix $\overline{\mathbf{G}}$ or the
control matrix $\mathbf{A}$ in (\ref{woronjetz}) may be singular or not
exist. Nevertheless, the DOA is a general concept, and input vector fields
can indeed be assigned to any (possible singular) configuration. Then the
DOA would change at input singularities. For instance a Gough-Stewart
platform is non-redundantly full-actuated as long as it does not enter the
well-known input singularities where the prismatic joint screws become
dependent and form a 5 system. In these input singularities the control
matrix $\mathbf{A}^T$ has rank 5. Hence the DOA reduces to 5, and the PKM is
redundantly underactuated in this input singularity.
\end{remark}

The geometric meaning of redundancy can vividly be explained for the
two-dimensional c-space $V\in \mathbb{R}^3$ in figure \ref{figScheme}.
Clearly, at the indicated c-space singularity, $V$ loses its manifold
property, as one can not assign a two-dimensional tangent plane at this
point. The two manifolds, separated by the c-space singularity, are the
motion modes of this fictitious PKM. The system has global and local DOF $%
\delta =2$, and can be locally controlled using joint variables $q^2$ and $%
q^3$ as inputs. Then, the PKM is non-redundantly full-actuated. The input
space $\mathcal{I}$ is the $q^2$-$q^3$ section of $V$, and $q^2$ and $q^3$
are local coordinates on $V$. This parameterization fails when the PKM
attains a configuration that projects to the boundary of $\mathcal{I}$. The
corresponding points in $V$ are the input singularities, as depicted in
figure \ref{figScheme}. The latter separate the c-space $V$ into the
indicated actuation modes. However, the PKM cannot be steered from actuation
mode 1 to 2 using the inputs $q^2$ and $q^3$. This would be possible using $%
q^1$ and $q^2$ with input space $\mathcal{I}^{\prime }$. Obviously, also for 
$\mathcal{I}^{\prime }$ there are input singularities, as indicated. Neither 
$\mathcal{I}$ nor $\mathcal{I}^{\prime }$ alone is a globally feasible input
space, but the combination of these (non-redundant) actuation schemes gives
rise to one with input space $\mathcal{I}\cup \mathcal{I}^{\prime }\equiv V$
and (redundant) inputs $q^1,q^2,q^3$, which is free of input singularities.
Then, the PKM is always redundantly full-actuated.

\begin{example}
%TCIMACRO{\TeXButton{\rm}{\rm} }
%BeginExpansion
\rm%
%EndExpansion
Consider two actuation schemes of the planar mechanism with DOF $\delta =2$
in figure \ref{fig:terminology}. First assume that joints 5 and 6 are
actuated. The configuration a) is an input and passive singularity, as the
motion of joints 1,3, and 8 are instantaneously independent from the input
motion. From there the mechanism can enter an actuation mode, of which a
configuration is shown in b). In this mode the DOA is $\alpha =2$, the
system is non-redundantly full-actuated. It can leave this actuation (and
motion) mode when steered into the c-space singularity c), where two
branches (motion modes) of the configuration space intersect. Figure \ref
{fig:terminology}d) shows a configuration in one of the possible actuation
modes with DOA $\alpha =1$, so that the mechanism is redundantly
underactuated with actuation redundancy $\rho =1$. In fact, the motion of
joints 1,3, and 8 can not be controlled by the actuated joints 5 and 6.%
\newline
Now assume that, in addition to joints 5,6, also joint 1 is actuated. Then,
configurations a) and b) belong to a single actuation mode, where the system
is redundantly full-actuated. That is, a) is not an input singularity for
this redundant actuation scheme. After passing through the c-space
singularity, in the actuation mode d) the mechanism is again redundantly
full-actuated. 
\begin{figure}[tbh]
\centerline{\includegraphics[width=8cm]{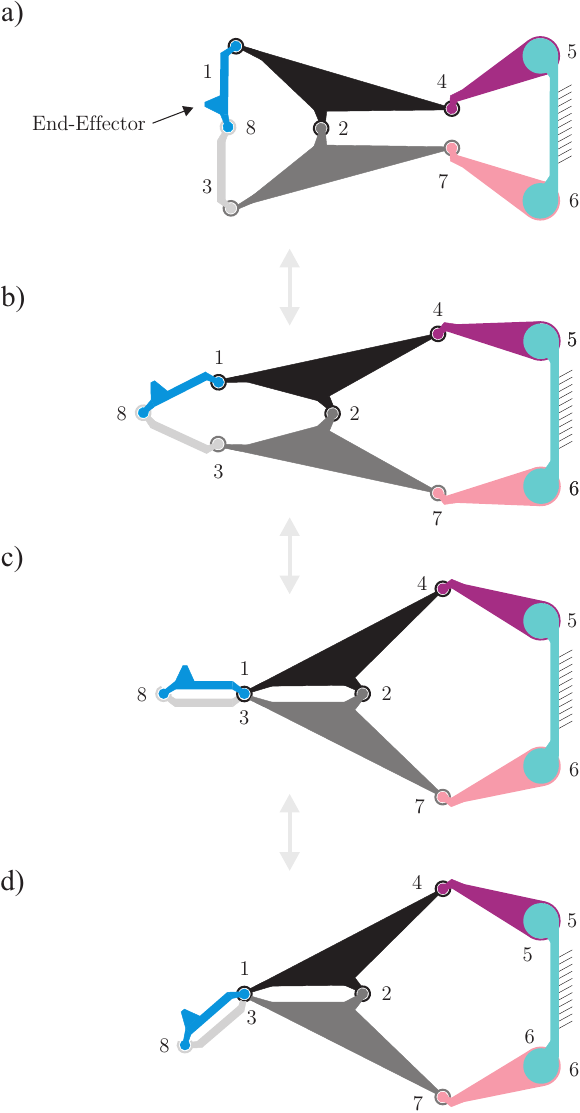}}
\caption{A 2 DOF manipulator in different actuation modes.}
\label{fig:terminology}
\end{figure}
%TCIMACRO{
%\TeXButton{\setcounter{example}{0}}{\setcounter{example}{0}
%}}
%BeginExpansion
\setcounter{example}{0}
%
%EndExpansion
\end{example}

\begin{example}[cont.]
%TCIMACRO{\TeXButton{\label{exam:pantograph3}}{\label{exam:pantograph3}}}
%BeginExpansion
\label{exam:pantograph3}%
%EndExpansion
%TCIMACRO{\TeXButton{\rm}{\rm} }
%BeginExpansion
\rm%
%EndExpansion
In the configuration $\mathbf{q}_0$ of the 5-bar mechanism in figure \ref
{figPantograph}a) the constraint and actuator Jacobian have full rank, $%
\mathrm{rank\,}\mathbf{J}\left( \mathbf{q}_0\right) =3$ and $\mathrm{rank\,}%
\mathbf{J}_{\text{a}}\left( \mathbf{q}_0\right) =2$, respectively, but $%
\mathrm{rank\,}\mathbf{J}_{\text{p}}\left( \mathbf{q}_0\right) =2$.
Moreover, since $\mathbf{J}_{\text{p}}$ has full rank 3 outside $\mathbf{q}%
_0 $ this is a passive singularity. There are possible instantaneous
motions, $\dot{\mathbf{q}}_{\text{p}}\in \ker \mathbf{J}_{\text{p}}$, of
passive joints without actuator motions. Due to $\mathrm{rank\,}\mathbf{J}%
\left( \mathbf{q}_0\right) >\mathrm{rank\,}\mathbf{J}_{\text{p}}\left( 
\mathbf{q}_0\right) $ not all actuator motions are feasible. In fact only
coupled actuator motions are possible in this configuration. This situation
is classified in \cite{zlatanovclassify} as a singularity of redundant
passive motion (RPM) and impossible input (II) type. If one would assign a
DOA to this configuration it would be $\alpha \left( \mathbf{q}_0\right) =0$%
. Otherwise the 5-bar mechanism has DOA $\alpha =2$. The 5-bar mechanism can
be regarded as instantaneously redundantly underactuated in $\mathbf{q}_0$
as the control forces cannot fully actuate the mechanism. Moreover, if the
mechanism is at rest in $\mathbf{q}_0$, and if there are no external or
inertia forces, it is not controllable. The images of the actuation modes in
workspace are visible in figure \ref{figManip}a). The two actuation modes
are separated by the input singularities.\newline
The redundantly actuated \underline{R}R/2\underline{R}RR PKM does not
possess such input singularities, and it has a single uninterrupted
actuation mode.
\end{example}

\begin{remark}
%TCIMACRO{\TeXButton{\rm}{\rm} }
%BeginExpansion
\rm%
%EndExpansion
Apparently actuation redundancy allows for elimination of input
singularities. A fundamental question is what degree of redundancy is
sufficient for achieving full actuation in all regular configurations. A
preliminary answer follows from the embedding theorem by Whitney \cite
{GuilleminPollack1974}, that, adapted to this problem, states that any $%
\delta $-dimensional c-space can be embedded in an Euclidean space of
dimension $2\delta +1$. It is not sure, however, that $2\delta +1$ actuators
are sufficient for full actuation. In case of the 5-bar linkage the
mechanism could always be fully actuated using 5 inputs. But one must be
careful, since the theorem only says that $V$ can embedded in some Euclidean
space. It does not say that this space is the (non-Euclidean) joint space.%
%TCIMACRO{
%\TeXButton{\setcounter{example}{2}}{\setcounter{example}{2}
%}}
%BeginExpansion
\setcounter{example}{2}
%
%EndExpansion
\end{remark}

\begin{figure}[t]
\centerline{\includegraphics[height=18cm]{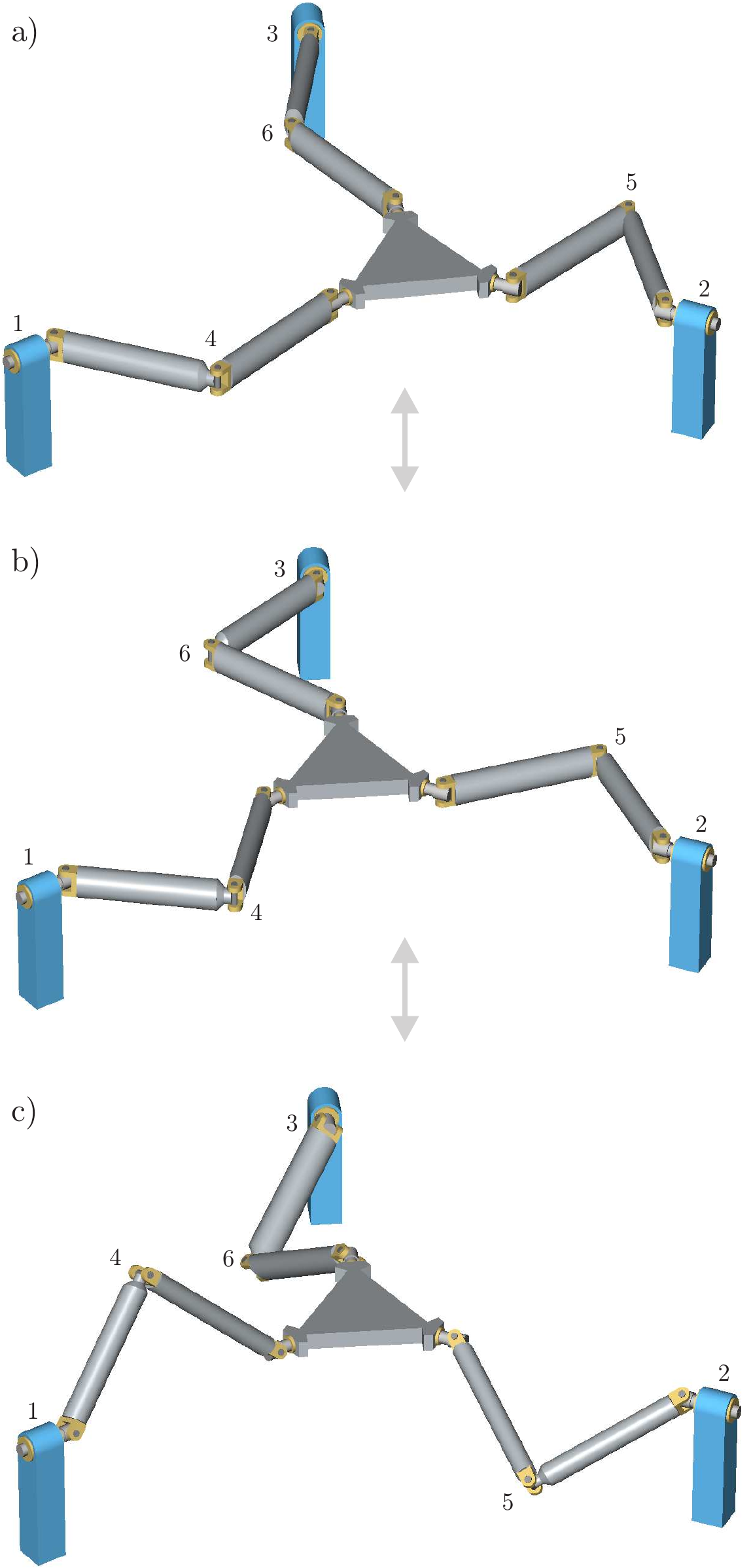}}
\caption{3URU DYMO PKM in its planar operation mode a), c-space singularity
b), and lockup mode c).}
\label{fig:3URUcomplete}
\end{figure}

\begin{example}
%TCIMACRO{\TeXButton{\rm}{\rm} }
%BeginExpansion
\rm%
%EndExpansion
Consider the 3-URU Double-Y Multi-Operational (DYMO) PKM in figure \ref
{fig:3URUcomplete}a) that was reported in \cite{zlatanovConstraintSing}.
This PKM, with mobility $\delta =3$, exhibits operation modes with different
degrees of actuations. First consider the actuation scheme with joints 4-6
actuated. In this mode the PKM acts as a planar manipulator since the
platform can only move in the horizontal plane. The PKM configuration is
completely determined by the active joints, so that in this mode the PKM is
non-redundantly full-actuated with DOA $\alpha =3$. It can leave this
operation mode via the c-space singularity in figure \ref{fig:3URUcomplete}%
b) where different motion modes (branches of the configuration space)
intersect. In this singularity the platform center is above the center of
the base triangle. Figure \ref{fig:3URUcomplete}c) shows a configuration in
one of the possible operation modes. This mode was called the lockup mode 
\cite{zlatanovConstraintSing} since the platform is immobile. In fact the
platform motion is independent of the motion of the limbs. With joints 4-6
actuated the PKM is redundantly underactuated with DOA $\alpha =0$ and
degree of actuation redundancy $\rho _\alpha =3$. That is, not only is the
platform fixed but the PKM motion cannot even be controlled as the limbs can
freely spin. In order to fully actuate the PKM in this mode one needs to
actuated joints 1-3. Now, if one intends to take advantage of the lockup
mode one would need additional (possibly low powered) actuators in joints
1-3. Then, with joints 1-6 actuated, the PKM is redundantly full-actuated
with DOF $\alpha =3$ and $\rho _\alpha =3$. Note, that this redundancy is
not achieved by the addition of kinematic chains connecting ground and
moving platform, but by the activation of passive joints.\newline
This redundant actuation would also be required to operate in the further
modes of this PKM that exhibits a spherical and mixed mode of the platform
motion \cite{zlatanovConstraintSing}. Using the redundant actuation scheme
with joints 1-6 the PKM is full-actuated in any mode.
\end{example}

\section{Summary}

In this paper a terminology for redundant PKM has been proposed. The aim of
this contribution was to provide consistent definitions upon a general
model, and to highlight the geometric aspects of redundancy. To this end a
kinematic model is introduced with the c-space as central part. Input,
output, and c-space singularities are distinguished and used to introduce
motion modes, actuation modes, and operation modes. The notion of kinematic
redundancy is recalled and task redundancy is introduced.

A dynamic model was introduced that enables to treat PKM as non-linear
control systems and to define actuation. The degree of actuation was
introduced as the number of independent control vector fields, and PKM are
classified as full-actuated and underactuated. Further, actuation redundancy
is defined as the difference of the number of actuators and the degree of
actuation. It is pointed out geometrically that input singularities can be
avoided by redundant actuation.

\end{document}

%% file: opening_PKMTerminology_revised.inp
\author{
	Andreas M\"uller
	\affiliation{ Chair of Mechanics and Robotics, University Duisburg-Essen\\
	Lotharstrasse 1, 47057 Duisburg, Germany\\
	andreas-mueller@uni-due.de 
	} 
\vspace{-6mm}
} 

\title{ 
\vspace{-7mm}
On the Terminology and Geometric Aspects of Redundant Parallel Manipulators
\vspace{-4mm}
} 
\maketitle 

\begin{abstract}
Parallel kinematics machines (PKM) can exhibit kinematic as well as
actuation redundancy. While the meaning of kinematic redundancy has been
clarified already for serial manipulators, actuation redundancy, that is
only possible in PKM, is differently classified in the literature. In this
paper a consistent terminology for general redundant PKM is proposed. A
kinematic model is introduced with the configuration space (c-space) as
central part. The notion of kinematic redundancy is recalled for PKM.
C-space, output, and input singularities are distinguished. The significance
of the c-space geometry is emphasized, and it is pointed out geometrically
that input singularities can be avoided by redundant actuation schemes. In
order to distinguish different actuation schemes of PKM a non-linear control
system is introduced whose dynamics evolves on the c-space. The degree of
actuation (DOA) is introduced as the number of independent control vector
fields, and PKM are classified as full-actuated and underactuated. Relating
this DOA to the degree of freedom (DOF) allows to classify the actuation
redundancy.

{\it Keywords--} {Parallel manipulator, terminology, kinematic redundancy, actuation redundancy, singularities}

\end{abstract}